**Style, Content, and the Success of Ideas**

Reihane Boghrati[1], Jonah Berger[2], and Grant Packard[3]

[1] Arizona State University

[2] University of Pennsylvania

[3] York University

**Abstract**

Why do some things succeed in the marketplace of ideas? While some argue that content drives success, others suggest that style, or the way ideas are presented, also plays an important role. To provide a stringent test of style's importance, we examine it in a context where content should be paramount: academic research. While scientists often see writing as a disinterested way to communicate unobstructed truth, a multi-method investigation indicates that writing style shapes impact. Separating style from content can be difficult as papers that tend to use certain language may also write about certain topics. Consequently, we focus on a unique class of words linked to style (i.e., function words such as "and," "the," and "on") that are completely devoid of content. Natural language processing of almost 30,000 articles from a range of disciplines finds that function words explain 13–27% of language's impact on citations. Ancillary analyses explore specific categories of function words to suggest how style matters, highlighting the role of writing simplicity, personal voice, and temporal perspective. Experiments further underscore the causal impact of style. The results suggest how to boost communication's impact and highlight the value of natural language processing for understanding the success of ideas.

Across different jobs, organizations, and industries, people want things to catch on. Politicians want their policies to be adopted and leaders want their strategies to be embraced. Health officials want their messages to diffuse and marketers want their products to succeed. Employees want their ideas to be approved and academics want their work to be cited.

Not surprisingly, then, across disciplines, researchers have long been interested in new product adoption, diffusion, and the success of ideas (Cosme et al., 2021; Heath et al., 2001; Kashima, 2000; Lieberson, 2000; Norenzayan et al., 2006; Rogers, 2010; Salganik et al., 2006; Schaller et al., 2002). Some innovations diffuse while others stagnate (Rogers, 2010). Some management practices are widely adopted while others languish (Rao et al., 2000). And some stories, stereotypes, products, and ideas take off while others never get traction (Fast et al., 2009; Schaller et al., 2002).

But why do some things succeed in the marketplace of ideas?

One possibility centers on content. Certain things are more successful because they are better or more novel than what came previously (e.g., relative advantage; Rogers, 2010). High speed internet replaced dial-up because it was faster and easier to use. Einstein's theory of general relativity replaced Newton's law of universal gravitation because it better explained the experimental evidence.

Another possibility, however, has less to do with content and more to do with style. This perspective suggests that the manner, or style, with which ideas are presented shapes their impact (Lakoff & Johnson, 1999). But does style actually shape the adoption of ideas, and if so, how?

To provide a particularly stringent test of style's importance, we examine it in a context where one might imagine content is paramount and style is not: academic research. Science prides itself on being an objective exercise, where writing is merely a disinterested way to communicate unobstructed truth (American Psychological Association, 2020; Pinker, 2014). The notion is that some discoveries (e.g., general relativity or prospect theory) are simply more novel or groundbreaking than others, and citations are seen as an unbiased measure of such quality (Hamermesh et al., 1982; Hamermesh & Schmidt, 2003; Smart & Waldfogel, 1996). Consequently, if presentation style matters even in such a content focused domain like academic research, it highlights its broader importance across domains.

Testing style's impact, however, is challenging. It's one thing to theorize that certain presentation styles are better, but actually measuring adherence to stylistic approaches and linking them to a consequential outcome is difficult.

Further, it can be tough to separate style from content. Even if papers that write certain ways (e.g., use more emotional language) are cited more, this could be driven by the subject matter. Papers studying certain topics (e.g., emotions) likely use more language related to those topics (i.e., emotional language), and thus content, rather than writing style itself, could be driving impact.

## Testing Style's Impact

To address these challenges, we focus on a small class of words that play a unique role in communication. Function words (e.g., conjunctions, grammatical articles, and prepositions, such as "and," "the," and "on") make up only a tiny portion of the human vocabulary (i.e., ~0.04%; Baayen et al., 1995) but appear in every sentence. They convey little semantic value on their own, but bind and enrich the nouns, adjectives, verbs, and some adverbs that make up communication content (Ireland & Pennebaker, 2010). The word "and," for example, could appear when people are writing about relationships (e.g., him and her), motivation (e.g., goals and motivation), or any number of content related topics. Because they are largely meaningless without content, function words are often treated as junk by language scholars (Chung & Pennebaker, 2007) and tossed out like meaningless garbage before text analysis is performed ("stop words," Lo et al., 2005).

But while function words tend to receive little attention from both scholars and communicators, they are particularly valuable because they capture communication style rather than content. As Chung & Pennebaker (2007) note, "the ways people use function words reflects their linguistic style" (p. 247). Consider two people saying that they like ice cream. Person A might say "I'd say that I really like eating ice cream" while Person B might say "Eating a scoop of ice cream is so enjoyable." While the content of what they are saying is the same (i.e., they both love ice cream), they express that same content using different linguistic styles. They use different function words (i.e., Person A uses personal pronouns and conjunctions while Person B uses articles and prepositions) to express their ideas.

Indeed, decades of research in linguistics, psychology, and other disciplines refer to function words as "style words" because they are seen as reflecting things about a communicator's linguistic style rather than anything about what is being discussed (Doré & Morris, 2018; Ireland et al., 2011; Pennebaker, 2011, 2017; Rosenthal & Yoon, 2011; Tausczik & Pennebaker, 2010). That is, function words reflect how people communicate while content words indicate what they are communicating (Tausczik & Pennebaker, 2010). Consequently, if how authors use different types of function words helps explain the impact of academic research, above and beyond any impact of content, it suggests that style matters.

To test this possibility, we examine tens of thousands of articles from a range of disciplines. First, we examine whether style matters. That is, even controlling for article content (e.g., its topical focus or research area), whether writing style (i.e., the function words used) is associated with the number of citations an article receives.[1] Given that factors beyond the article text also likely drive citations (e.g., journal, publication year, and author prominence), we test whether the observed effects persist even including a range of controls. To allow for non-linear relationships, or interactions between variables, we also examine whether the results hold using a machine learning model optimized for prediction.

Second, we provide an exploratory analysis of how style matters. That is, whether particular ways of writing are associated with greater impact. To ensure the language used is capturing style rather than content, we again focus on function words, exploring whether, controlling for article content, particular ways of using functional words are associated with greater impact.

Third, we conduct some simple follow-up experiments to underscore the causal impact of some of the features identified in the exploratory analysis.

Note, we do not mean to suggest that this is the first paper to examine the link between writing and citations. Indeed, individual papers in different disciplines have explored such relationships (e.g., August et al., 2020; Freeling et al., 2019; Stremersch et al., 2007).

Prior work, however, suffers from a key issue. While it documents correlations between language features and citations, it is often unclear whether any observed effects are driven by (1) language itself versus (2) differences in correlated *content*, or the types of academic papers that tend to use different types of language. Warren et al. (2021), for example, suggests that papers that use more technical or abstract language are cited less, but these relationships could be driven by content rather than style. Papers that use more abstract (e.g., "subtle fault") or technical language (e.g., "endogeneity") are often about completely different topics (i.e., different research areas) than those that use less abstract or technical language, so the difference in citations could be driven by the content itself. In sum, by focusing mainly on content words (i.e., adjectives, nouns, and verbs) prior work often confounds style with content.

---

[1] Note that we are not suggesting that using *more* function words is necessarily better. Rather, consistent with prior work, we simply suggest that the function words communicators use reflects their linguistic style, and this, in turn, may shape the impact their ideas receive. This may involve more of certain function word categories (e.g., conjunctions) and less of others (e.g., prepositions), which we examine later.

We address this issue three ways. First, we specifically examine function words because they are the only language features that allow one to unconfound style from content. Function words (e.g., "the" or "an") are unrelated to an article's content and thus provide a unique identification strategy to separate content and style. If language features that have no relation to article content are still linked to citations, it shows that style matters even above and beyond content. Thus, we not only look at different language characteristics than prior work, we examine the only language features that allow one to isolate the role of style in cultural success.

Second, to further demonstrate the effect cannot be driven by content we control for article content using topic modeling. We represent each article as a mixture of different topics, and control for this mix across articles. This has not been done in prior work and further separates the effects of content and style.

Third, to provide an even stronger test of causality, we conduct simple experiments. While we examine language features devoid of content, and control for content using topic modeling, one could still wonder whether there is some unobserved factor driving both the use of function words and citations. Note that controlling for things like author fame, institution, and various other linguistic features reduces the possibility of such an unobservable factor, but to offer an even more direct test, we experimentally manipulate some of the language features identified and measure their impact.

We are not suggesting function words are the only linguistic factor that may affect citations, or even the ones with the largest impact. Rather, we pick this category because it is the only one that allows the separation of style from content. Beyond simply looking at a different set of linguistics characteristics, one key contribution of this work is causally distinguishing the impact of content and style. Further, by testing this across tens of thousands of articles across several different disciplines, we demonstrate the generalizability of the effect.

## Does Style Matter?

**Method**

We compiled a corpus of full-text peer-reviewed articles from 1990–2018 from five social science disciplines (i.e., psychology, economics, political science, anthropology, and sociology). Online rankings and discipline experts were used to identify the top journals (e.g., Econometrica, Psychological Science, and American Journal of Sociology) in each discipline. Then we acquired full-text article data, as well as meta-information (e.g., title, issue, and authors), from as many of these journals as possible from JSTOR, a digital library of academic journals (see Table A1 for a summary of data).

Given the focus on research articles, we removed non-research articles (e.g., letters to the editor) and articles with missing text. To remove non-research articles, we searched the meta-information for articles whose subjects included words such as "reports," "comment reply," or "note" and searched for phrases like "letter to the editor," "front matter," or "in this issue" in article titles. In case this missed some non-research articles, we also removed articles with extremely small word counts, calculating mean word count per journal and removing articles three standard deviations below the mean. JSTOR automatically exports PDFs to text files, and some articles were transferred incorrectly, so we removed any article whose word count per page fell below 200 words per page. This left a dataset of 28,774 articles.

Given the focus on writing style of an article, rather than the titles of articles it referenced, references were removed. We identified the last occurrence of the word "references" and calculated the concentration of four-digit numbers beginning with "20" and "19" (as calendar year references to the current or prior century). If those numbers occurred more than a reasonable portion of the time (i.e., 0.5%, or one out of every 40 words, which is around the number of words in each reference), we removed whatever came after.

To capture writing style, following prior work (Chung & Pennebaker, 2007; Ireland & Pennebaker, 2010) we measured the incidence rate (proportion of words) of each of the nine categories of function words (i.e., conjunctions, prepositions, quantifiers, negations, grammatical articles, personal pronouns, impersonal pronouns, auxiliary verbs, and common adverbs) in each article measured through Linguistic Inquiry and Word Count (LIWC; Pennebaker et al., 2015). As noted, we use function words in particular because they are devoid of content and thus can separate out the effect of style from content.

To measure the dependent variable, we collected the number of citations each article received from Google Scholar using the Publish or Perish program (Harzing, 2007).

**Controls**

Citations are obviously driven by many things beyond the text of an article, so to test alternative explanations and robustness, we control for a variety of other factors.

First, we control for various non-language features that might impact citations. Articles in certain journals may receive more citations, so we included dummy variables to control for journal. Note that this also controls for any general differences in citation rates across disciplines. Older articles should have more opportunity to accrue citations, so we control for publication year. We also control for a number of features shown in prior work to relate to citations, including article length (Stremersch et al., 2007), abstract length and title length (Annalingam et al., 2014), article order in the journal (Stremersch et al., 2007), number of authors[2] (Chen, 2012), and number of references (Chen, 2012).

To control for author gender, we relied on methods from prior research (Fox et al., 2016; Topaz & Sen, 2016). Genderize.io uses public census data to map each author name to "male," "female," or "none." This combined with the gender_guesser python package (version 0.4.0) and the national name list (Kaggle, 2017) successfully extracted authors' gender for most of the articles. For the remaining authors, research assistants manually identified whether an author was female or male by searching for them on the internet.

Certain types of articles (e.g., theory pieces) may be cited more, so we use supervised machine learning to classify articles as theory pieces or not. Four research assistants manually coded 700 articles based on whether they included any experiments, data analysis, or an analytical model. Pieces that did not do any of these things were classified as theory pieces. This served as a training set. Next, the feature vector for the machine learning approach was generated, including the presence of equations (calculated by the percentage of equal signs in an article), empirical topics (calculated using LDA to identify 17 empirical topics and summing up the topic probabilities for each article), and tf-idf features based on articles' main text. Support vector machines (SVM) were then applied on the training dataset, achieving 91% accuracy using 10-fold cross-validation. Finally, the trained model identified the article label (i.e., theory or not) for the rest of the articles.

---

[2] While one might suggest controlling for authors, unfortunately this was not feasible. Very few author groups appear more than once in the data and the model did not converge when author random or fixed effects were included.

Second, we control for article content. Certain research topics or areas (e.g., trade policy or self and identity) might be cited more (Willis et al., 2011), so we control for this using a well-adopted topic modeling method, Latent Dirichlet Allocation (LDA; Blei et al., 2003). Rather than assuming each article covers only one theme, topic modeling allows each article to be represented as a mixture of different topics (e.g., 10% topic A and 7% topic B). Based on coherence scores an 80-topic approach was used. We included each article's topic probabilities as controls in the model. Consistent with the notion that function words are capturing style and not content, their average absolute correlation with topics is only 0.05.

**Main Analytic Approach**

Given citations are counts, and over-dispersed, we used a negative binomial regression to analyze the link between style features and citation count. Because different models use different numbers of predictors, we use adjusted $R^2$ for model comparison, which adjusts variance explained for the number of predictors in the model. Results are identical if out-of-sample cross-validation is used for model comparison instead (see Table A2). As reported in the Supplemental Materials, for this robustness check, we use 10-fold cross-validation and in each iteration, train a model with 90% of the data (training set), apply it against the remaining 10% of the data (test set), and calculate the predicted $R^2$ on the test set.

**Results**

Results suggest that above and beyond the variance explained by non-language features (model 1, *adj. $R^2$* = 0.129), adding style features helps explain how many citations articles receive (model 2, *adj. $R^2$* = 0.149, *F* = 16.69, *p* < .001, see Table 1). Adding style features also adds predictive power even once article content is included (model 3 vs. model 4, *adj. $R^2$* = 0.217 vs. 0.208, *F* = 2.665, *p* = .004, see Table 1). Results also persist controlling for author prominence (i.e., author fame or prestige of institution), where authors are from (i.e., within or outside the United States), and other content controls (see Robustness Checks).

**Table 1: Style Words and Citations**

|  | Negative Binomial | | | | Machine Learning | |
|---|---|---|---|---|---|---|
|  | (1) Baseline | (2) +Style | (3) Baseline+Content | (4) +Style | (5) Baseline+Content | (6) +Style |
| **Adjusted $R^2$** | 0.129 | 0.149*** | 0.208 | 0.217** | 0.209 | 0.234*** |
| Style Features |  | yes |  | yes |  | yes |
| Content Controls |  |  |  |  |  |  |
|   LDA Topics |  |  | yes | yes | yes | yes |
| Non-Language Controls |  |  |  |  |  |  |
|   Publication Year | yes | yes | yes | yes | yes | yes |
|   Journal | yes | yes | yes | yes | yes | yes |
|   Article Length | yes | yes | yes | yes | yes | yes |
|   Abstract Length | yes | yes | yes | yes | yes | yes |
|   Title Length | yes | yes | yes | yes | yes | yes |
|   Article Order | yes | yes | yes | yes | yes | yes |
|   Num Authors | yes | yes | yes | yes | yes | yes |
|   Author Gender | yes | yes | yes | yes | yes | yes |
|   Num References | yes | yes | yes | yes | yes | yes |
|   Article Type | yes | yes | yes | yes | yes | yes |
| Observations | 28774 | 28774 | 28774 | 28774 | 28774 | 28774 |

*Note.* ****p* < .001, ***p* < .01, **p* < .05, values compared adding style features in each of the three model comparisons.

These results are intriguing, but one could wonder whether they might somehow be driven by the modeling approach used. While results are the same using a penalized LASSO regression (Tibshirani, 1996) to identify and remove predictors that may introduce collinearity (*adj. $R^2$* = 0.217 vs. 0.209, $F$ = 2.658, $p$ = .004), maybe there are non-linear relationships between the non-language or content controls and citations, or interactions between these variables that, once included, would wipe out any effect of style features.

To test this possibility, we also used a machine learning model (i.e., a two-layer feed-forward neural network). This approach is useful because it tries all combinations of variables to achieve the best prediction. Consequently, it should improve the baseline prediction of the model without style features, allowing for a stronger test of whether style matters. We implemented the neural network using the Keras Python package (version 2.4.3), with two hidden layers with 64 and 32 nodes and TanH as the activation function. Given that the output is count data, the output layer's activation function was set to exponential. As the data comes from the Poisson distribution, a Poisson loss function was used. Even using this more sophisticated approach, however, including style features still adds additional predictive power (model 5 vs. model 6, *predicted $R^2$* = 0.234 vs. 0.209, $F$ = 7.947, $p$ < .001).

Overall, these results suggest that function words, which make up less than 1% of unique words used, explain 13–27% of language's impact on citations

**Robustness Checks**

To further test alternative explanations, we conducted a number of robustness checks.

***Other Content Features.*** First, one could wonder whether the model has truly controlled for content. To further address this possibility, we included the prevalence of the psychological process dictionaries from Linguistic Inquiry and Word Count (i.e., affective processes, social processes, cognitive processes, perceptual processes, biological processes, drives, time orientations, relativity, personal concerns, and informal language; Pennebaker et al., 2015) which include over 30 sub-dictionaries (e.g., topics such as family, health, and religion). Results remain the same: including style features still adds predictive power (*adj. $R^2$* = 0.220 vs. 0.211, *F* = 2.382, *p* = .011). Results also remain the same controlling for standard measures of readability (e.g., Flesch reading ease, *adj. $R^2$* = 0.217 vs. 0.208, *F* = 2.794, *p* = .003).

***Where Authors Are From.*** Second, one could wonder whether the results are driven by where authors are from. Scholars outside the U.S. may write differently or be cited less (Stremersch & Verhoef, 2005). To test whether this can explain the results, we collected data on author institutions from Elsevier. We identified which institution each author was associated with, and whether the institution was based in the U.S. or not. For each academic article, we then calculated the percentage of authors that were U.S. based and used that as a control variable. While this measure was only available for a portion of the papers in the dataset, even controlling for whether institutions were outside the U.S., including style features still adds predictive power (*adj. $R^2$* = 0.238 vs. 0.231, *F* = 1.885, *p* = .049).

***Author Prominence.*** Third, one might wonder whether the results are driven by author prominence. Maybe more prominent authors use particular writing styles, and it is their prominence, rather than the writing style itself, that impacts citations.

This does not seem to be the case. To measure author prominence, we first scraped authors' Google Scholar profiles to retrieve their total citation count. This was only available for 57% of authors in the dataset, so to retain as many articles as possible, we used maximum citation count among all authors of a given article as the measure of prominence. Even including author prominence in the model, including style features still adds predictive power (*adj. $R^2$* = 0.168 vs. 0.100, *F* = 15.440, *p* < .001).

We also measured prominence through authors' academic affiliations. We matched author institution data with global institution rankings from timeshighereducation.com and averaged school rankings over the past 10 years. For authors whose school rank was not available, the school rank was set to the lowest rank plus one (i.e., 201). Again, author affiliation was only available for a portion of authors in the dataset, so to retain as many articles as possible we used the highest-ranking school from the authors of a given paper as a measure of prominence. Results remain the same (*adj. $R^2$* = 0.228 vs. 0.221, *F* = 2.249, *p* = .017).

Overall then, while author prominence certainly helps explain citations, and prominent authors may even write differently, function words still explain citations even controlling for these aspects. Further, even controlling for all the robustness check factors in one model, function words still add predictive power (Table A3).

*Abstracts*. Fourth, though not an alternative explanation, one might wonder whether abstracts alone are enough to explain the effects. If readers were only considering abstracts, this should decrease the ability to find a relationship between article language and citations, as readers would not be exposed to most of the language in the article.

That said, to test this possibility, we extracted the abstract from each article. The abstracts' texts are available in meta-information files, but they don't always appear the same way in the articles' text (e.g., different spellings are used, or they are mixed up with author information). Consequently, we cannot search for exact matches and the following approach was taken. First, each abstract's text was extracted from the meta-information files. Second, both the abstract's text and the article's main text were tokenized into sentences. Third, starting from the beginning of main text, the first text window with the same length as the abstract was compared to the abstract, and if the similarity was above 90%, the text window was labeled as the abstract and removed from the main text. If not, the window was shifted by one sentence and the analysis repeated until the abstract was found or the window was shifted 20 sentences, whichever came first. Similarity was calculated using the FuzzyWuzzy Python package (version 0.17), which uses an edit distance algorithm.

While abstracts were only available for 84% of articles, results indicate that compared to a base model including all the controls, adding style features for only the abstract does not increase predictive power (*adj. $R^2$* = 0.218 vs. 0.217, *F* = 0.259, *p* = .985). This casts doubt on the notion that the effects are driven solely by abstracts.

**Discussion**

Taken together, the results suggest that style matters. Incorporating style features (i.e., how authors use different types of function words) increases the variance explained by 1.0–2.5%, which is 4–11% of the overall variance explained and 13–27% of the variance explained by language content (i.e., LDA topics). This result suggests that fewer than 500 style words that contain no ideas or content on their own hold up against over 100,000 content words in explaining an idea's success.

## Writing Styles Associated with Impact

While the main results suggest that style matters, they bring up an important question: *how* does style matter? That is, which ways of writing increase impact? Exploratory analyses on function word categories suggest that personal voice, temporal perspective, and simplicity all play a role.

**Method**

Different writing styles might have different effects across different parts of an article. As examined below, for example, personal voice may be particularly important when introducing ideas (i.e., in the introduction of an article), but less important in the methods section. Consequently, we use manual coding and machine learning to separate each article into three segments, (1) front end (i.e., literature review and theorizing), (2) methods and results, and (3) general discussion or conclusion (see Supplemental Materials for supervised machine learning approach). Then, we use the negative binomial model from the main analyses including all the control variables (i.e., non-language and content controls) to test the relationship between function word categories and citations across different article segments (controlling for all other function word categories in each of those segments). This sheds light on the relationship between each language feature in a given article portion (e.g., articles in the front end) and citations, controlling for the presence of that feature in other portions (i.e., articles in the conclusion).

**Results**

*Personal Voice.* Academic writing guides have argued that authors should write in a manner that is distant, objective, and devoid of self-reference (e.g., first-person pronouns like "I" or "we"; Bem, 2003; Strunk Jr & White, 1999). But is that actually more effective?

In contrast to prior suggestions, results suggest that personal voice may sometimes be beneficial. Papers whose front ends are written with more self-referencing function words used (i.e., first person pronouns) are cited more ($b = 0.083$, $p < .001$). Papers written with more first-person pronouns in the middle, usually empirical section, however, are cited less ($b = -0.033$, $p < .05$).

This may reflect the degree to which personal ownership is valuable in different parts of a paper. Taking personal ownership of arguments, hypotheses, and contributions (e.g., "we reveal" vs. "the present research reveals") that are empirically supported may make the authors seem more prescient, increasing the perceived authority of the research. Taking personal ownership of methods and results (e.g., "we asked participants to do X" vs. "participants did X"), however, may make methodological choices seem more subjective. Along these lines, in the middle section, impersonal pronouns (e.g., "it" or "that"), which remove a personal actor from the methods and results section were associated with being cited more ($b = 0.077, p < .001$).

Looking beyond first-person pronouns, a multiple function word measure used to identify more personable writing style (i.e., analytic thinking; Jordan et al., 2019) shows similar results.

***Temporal Perspective***. Journal style guides, and academics themselves, commonly recommend describing research using the past tense (American Psychological Association, 2020; Bem, 2003; Nature, 2014). But is that actually more effective?

Analysis of a function word category (i.e., auxiliary verbs), as well as temporal language more generally, suggests past-focused language may actually reduce citations. Auxiliary verbs (e.g., "had" or "will") modify content verbs and can signal their framing in time (e.g., "had considered" or "will consider"). Judges manually coded each auxiliary verb based on whether it referenced the past, present, or future. Results indicate that while papers written with more past-focused auxiliary verbs are cited less ($b = -0.100, p < .001$), those written with more present-focused auxiliary verbs are cited more ($b = 0.072, p < .01$). LIWC's temporal orientation categories (i.e., focus past and focus present; Pennebaker et al., 2015) show similar effects. More past focused articles are cited less ($b = -0.118, p < .001$), while more present focused articles are cited more ($b = 0.081, p < .001$). While a paper's content (i.e., theorizing, methods, and results) occurred in the past, using present tense may make that content seem more current and in the moment (Liberman et al., 2007), which might make it seem more relevant, applicable, and important.

***Simplicity.*** While ideas are often quite complex, communicating them more simply might increase their impact. We propose this may be particularly important when ideas are first being explained (e.g., in the beginning of an article where authors are laying out their thinking and explaining how their work relates to prior research).

Consistent with this possibility, results indicate that papers whose front ends use less complex writing are cited more. *Articles* and *prepositions* are two types of style words linked to cognitive complexity (Biber et al., 2011; Pennebaker et al., 2003; Pennebaker & King, 1999). Grammatical articles ask readers to make distinctions between a single case or a class of something (e.g., *the* car vs. *a* car) while prepositions describe the nature of linkages between nouns, pronouns, or phrases (e.g., growth *despite* inflation; ate more candies *except* when). Consistent with the notion that less complex writing boosts citations, papers that use fewer articles ($b = -0.030$, $p < .001$) or prepositions ($b = -0.016$, $p < .0$) in the front end are cited more. Traditional readability measures show similar effects (Supplemental Materials).

While we are not the first to suggest simple writing can be good (e.g., Clayton, 2015; Day & Gastel, 2012) by using function words, we provide a stronger empirical test that simple writing itself, rather than different associated content (i.e., articles about simpler topics) is driving the effect.

Further, by separating different sections, our approach also allows us to ask whether simplicity is *always* good. In the methods and results, for example, complexity may sometimes be useful or even required. Indeed, the cost of complexity seems to weaken ($b_{articles} = -0.009$, $p = $

.10) or even reverses ($b_{prepositions} = 0.029$, $p < .001$) in the middle section where methods or results are discussed.

## Experimental Evidence

While natural language processing of tens of thousands of articles suggests that linguistic style shapes impact, one could still wonder whether the effects are truly causal. After all, maybe there is some unobserved factor that is driving both linguistic style and citations. Controlling for things like author fame, institution, article topics, other linguistic features, and over a dozen other measures casts doubt on this possibility, but to provide an even stronger causal test, we conduct two simple experiments. In each, we manipulate one of the styles linked to increased citations (i.e., personal voice or temporal perspective), and test its causal impact.

### Experiment on Personal Voice

*Method.* Participants ($N = 172$, recruited through Amazon Mechanical Turk) were randomly assigned to condition (i.e., Control or Personal Voice) in a between subjects design. They were told they would read some information about academic research and answer some questions that followed.

All participants read a research description, and the only difference between conditions was whether or not the description included personal voice. In the control condition, participants read that "Results reveal an even more cost-efficient way to produce clean energy." The personal voice condition was almost identical expect that a personal pronoun replaced one word so that it read "We reveal an even more cost-efficient way to produce clean energy."

All participants then completed the dependent measures, the importance of the research (i.e., "how important is this research") and its likely impact (i.e., "how impactful do you think this research will be"), both on 7-point scales (1 = not at all, 7 = extremely).

*Results*. As predicted, and consistent with the results of the field data, using personal voice led people to think the study was more important ($M_{\text{Personal Voice}} = 6.20$ vs. $M_{\text{Control}} = 5.86$, $F(1,170) = 5.59$, $p = .019$). Using personal voice also led them to think the study would be more impactful ($M_{\text{Personal Voice}} = 5.95$ vs. $M_{\text{Control}} = 5.42$, $F(1,170) = 7.46$, $p = .007$).

**Experiment on Temporal Perspective**

*Method.* Participants ($N = 171$, recruited through Prolific) were randomly assigned to condition (i.e., Present Tense or Past Tense) in a between subjects design. They were told they would read some information about academic research and answer some questions that followed.

All participants read a research description, and the only difference between conditions was the verb tense used. In the past tense condition, participants read that "This drug lowered the risk of cancer by 15%." The present tense condition was almost identical expect that the verb "lower" was shifted to the present tense so that it read "This drug lowers the risk of cancer by 15%."

All participants then completed the dependent measures, the importance of the research (i.e., "how important is this research") and its likely impact (i.e., "how impactful do you think this research will be"), both on 7-point scales (1 = not at all, 7 = extremely).

*Results.* As predicted, and consistent with the results of the field data, using present tense led people to think the study was more important ($M_{\text{Present}} = 6.06$ vs. $M_{\text{Past}} = 5.63$, $F(170) = 6.84$, $p = .010$). Using present tense also led them to think the study would be more impactful ($M_{Present} = 5.97$ vs. $M_{Past} = 5.54$, $F(170) = 6.07$, $p = .015$).

**Discussion**

Experimental evidence underscores the causal impact of style. Consistent with the field data, small shifts in the use of function words increased the perceived importance and expected impact of academic research. By directly manipulating these language features through simple experiments, we rule out alternative explanations based on unobservables or correlated content, and underscore function words' causal impact. By keeping the content identical, and simply manipulating style, we further demonstrate style's importance in the success of ideas.

**General Discussion**

Academics and practitioners alike have long debated about why things succeed in the marketplace of ideas. Some products, services, and ideas catch on while others languish. Why?

While content certainly matters, the present work suggests that style also plays an important role. Even in academic research, where writing is often seen as merely a disinterested way to communicate truth, writing style helps explain the impact (i.e., number of citations) ideas achieve. Style words (i.e., "function words"), which make up less than 1% of unique words used, explain 13–27% of language's impact on citations. If communication style shapes impact even in a content focused domain like academic research, its effects in other domains is likely even larger.

The results also suggest particular writing styles that may increase impact. While some have suggested that first person voice is bad because it distracts from the paper's content (Bem, 2003), results suggest that there are times for taking personal credit for writing (i.e., front end) and others for letting the paper's content stand on its own (i.e., methods and results). While journal style guides often suggest using past tense, results suggest that using the present tense (e.g., we *theorize* instead of *theorized*) may be more beneficial. And while academic ideas are often complex, explaining things simply may be important, particularly in a paper's front end.

**Contributions**

This work makes three main contributions. First, our empirical approach allows us to disentangle style from content and isolate whether style matters. Numerous language features might be correlated with citations, but in many cases, it's not clear whether those relationships are truly driven by the language feature itself versus differences in correlated content, or the fact that papers discussing certain types of content might be cited more. To address these issues, we focus on a unique class of words (i.e., function words) that are completely devoid of content. To further cast doubt on the role of content, we directly control for it (i.e., through topic modeling and other dictionaries) as well as other factors that could drive citations (e.g., author characteristics). Using a machine learning model that is more optimized for prediction also ensures that the results are not driven by assumptions about linear relationship or relationships between variables. Experiments further demonstrate the causal impact of style, even when the content is kept identical.

Second, we begin to examine how style matters, or which style features increase impact. We document a number of novel relationships and underscore their causal effect through experiments.

Third, we suggest *when* particular language features might matter more. Content is often divided into multiple parts, or portions. The goals of a business presentation's introduction, for example, might be quite different than the goals of the main part. (e.g., to garner interest in the topic versus inform). Similarly, the introduction of an academic paper may have different goals than the methods and results. Consequently, simple suggestions like "write simply" may not only miss nuance but be detrimental in some cases. Indeed, our results suggest that writing too simply in the methods and results may decrease impact. By dividing papers into sections, we begin to explore not only what matters but when. Hopefully future research can examine these aspects in greater detail.

**Implications**

These findings have clear implications. Most narrowly, peer-reviewed research often adopts a dry, dense, and impersonal style that can be sometimes challenging to read and understand (Freeling et al., 2019; Ruben, 2016). But while authors across many disciplines have intermittently theorized about what counts as "better writing" (Bem, 2003; McCloskey, 1985; Mensh & Kording, 2017), little work has actually tested these suggestions.

The present work suggests that a few relatively simple writing shifts may help boost an idea's impact. While writing guides often recommend writing in a distant, objective manner that avoids self-reference, using first-person pronouns (e.g., "we" suggest) in the front end of a paper may increase impact. Further, while style guides also commonly recommend past tense, this may be misguided. Using present tense rather than past (e.g., "we find" rather than "we found") may increase impact by making the ideas seem more current and relevant. And using a simpler style (i.e., fewer articles and prepositions), for example, particularly in the front end of the paper, may boost impact.

More broadly, the findings highlight the importance of communication style in the marketplace of ideas. Language plays an integral role in persuasion. Health officials use language to persuade citizens, politicians use language to persuade voters, and leaders use language to persuade employees. But as these results suggest, ideas succeed not only based on their content, but how they are framed and conveyed.

Consequently, when trying to get ideas to catch on, subtle linguistic shifts may increase impact. Communicators often think that using complex language will make them seem more intelligent (Oppenheimer, 2006), but simpler language may actually be more effective. Presenters might also benefit from considering which part of ideas to take ownership of. While owning arguments, directions, or contributions may be beneficial, when discussing the evidence for one's ideas, avoiding self-reference may make choices seem less subjective. Present tense may also increase persuasion. Saying a movie "is" good, for example, rather than "was" good may make others more interested in watching it. Similarly, online reviews that note that a restaurant "has" great food, rather than "had" great food, may be more helpful and encourage readers to go.

**Future Research**

Future work might examine the relative impact of style versus content in different domains. One could argue that academic writing is relative constrained. Papers have norms or style guides, and given the topic matter, there may be constraints on things like pronoun use or certain types of grammatical articles. Consequently, the effects might be even larger in areas like advertising or public speaking where the language is less constrained.

Future research might also delve into other ways language shapes impact. Expressing more certainty, for example, may be beneficial (because it increases the perception that a phenomenon is true) or detrimental (if it seems unwarranted; Willis et al., 2011). Using more familiar language may help if it makes things easier to read. Compared to using definite articles (i.e., ``the'', which specifies a singular, identified member of the type, e.g., violating *the norm of* …) using indefinite articles (i.e., "a" or "an" which means that any member of that type is being discussed, e.g., violating *a norm of* …), may make content seem broader and generally applicable, which may increase citations. Language's impact may also vary by discipline. While jargon may generally decrease readability (Pennebaker et al., 2015), and thus citations, it may increase impact in disciplines where it is seen as a signal of credibility.

Finally, this work highlights the value of using natural language processing to understand both human behavior and why things catch on. Almost everything involves language in one way or another. People produce language through writing emails, making presentations, talking to friends, and posting content online. Further, they consume language through watching movies, listening to songs, and reading books, news articles, and other types of content. Even human thought involves language.

But while natural language processing itself is not new, novel methods and data availability have opened up a range of exciting avenues for research. Researchers have examined things like gender bias, narrative structure, political divides, creativity, cultural success, and the universal structure of emotion (Berger & Packard, 2018; Boyd et al., 2020; Boyd & Pennebaker, 2017; DeFranza et al., 2020; Gray et al., 2019; Heath & DeVoe, 2005; Hsu et al., 2021; Jackson et al., 2019; Kesebir, 2017; Sterling et al., 2020; see Berger & Packard, 2021; Boyd & Schwartz, 2021 for recent reviews).

As these examples highlight, language serves two key functions. It both (1) *reflects* things about the people, organizations, and socio-cultural contexts that produce it, and (2) *impacts* the people and audiences that consume it. Consequently, analyzing language can shed light on both language producers and language recipients. It can deepen understanding around differences between people, organizations, and cultures. And it can provide insight into persuasion, effective communication, and why things catch on. By quantifying features of language, automated text analysis will hopefully unlock a range of interesting insights.

STYLE, CONTENT, AND THE SUCCESS OF IDEAS 29narrative structures through text analysis. *Science Advances*, *6*(32), eaba2196.

Boyd, R. L., & Pennebaker, J. W. (2017). Language-based personality: a new approach to personality in a digital world. *Current Opinion in Behavioral Sciences*, *18*, 63–68.

Boyd, R. L., & Schwartz, H. A. (2021). Natural language analysis and the psychology of verbal behavior: The past, present, and future states of the field. *Journal of Language and Social Psychology*, *40*(1), 21–41.

Chen, C. (2012). Predictive effects of structural variation on citation counts. *Journal of the American Society for Information Science and Technology*, *63*(3), 431–449.

Chung, C., & Pennebaker, J. W. (2007). The psychological functions of function words. *Social Communication*, *1*, 343–359.

Clayton, V. (2015). *The Needless Complexity of Academic Writing*. The Atlantic: Education. https://www.theatlantic.com/education/archive/2015/10/complex-academic-writing/412255

Cosme, D., Scholz, C., Chan, H.-Y., Doré, B. P., Pandey, P., Tartak, J. C., Cooper, N., Burns, S., Paul, A., & Falk, E. (2021). *Message self and social relevance increases intentions to share content: Correlational and causal evidence from six studies*.

Day, R. A., & Gastel, B. (2012). *How to write and publish a scientific paper*. Cambridge University Press.

DeFranza, D., Mishra, H., & Mishra, A. (2020). How language shapes prejudice against women: An examination across 45 world languages. *Journal of Personality and Social Psychology*.

Doré, B. P., & Morris, R. R. (2018). Linguistic synchrony predicts the immediate and lasting impact of text-based emotional support. *Psychological Science*, *29*(10), 1716–1723.

Fast, N. J., Heath, C., & Wu, G. (2009). Common ground and cultural prominence: How conversation reinforces culture. *Psychological Science*, *20*(7), 904–911.

STYLE, CONTENT, AND THE SUCCESS OF IDEAS                                          33difference. *Journal of Personality and Social Psychology*, *77*(6), 1296.

Pennebaker, J. W., Mehl, M. R., & Niederhoffer, K. G. (2003). Psychological aspects of natural language use: Our words, our selves. *Annual Review of Psychology*, *54*(1), 547–577.

Pinker, S. (2014). *Why academics stink at writing*. The Chronicle of Higher Education. https://www.chronicle.com/article/why-academics-stink-at-writing/

Rao, H., Morrill, C., & Zald, M. N. (2000). Power plays: How social movements and collective action create new organizational forms. *Research in Organizational Behavior*, *22*, 237–281.

Rogers, E. M. (2010). *Diffusion of innovations*. Simon and Schuster.

Rosenthal, J. S., & Yoon, A. H. (2011). Detecting multiple authorship of United States Supreme Court legal decisions using function words. *The Annals of Applied Statistics*, 283–308.

Ruben, A. (2016). *How to read a scientific paper*. Science. https://www.sciencemag.org/careers/2016/01/how-read-scientific-paper

Salganik, M. J., Dodds, P. S., & Watts, D. J. (2006). Experimental study of inequality and unpredictability in an artificial cultural market. *Science*, *311*(5762), 854–856.

Schaller, M., Conway III, L. G., & Tanchuk, T. L. (2002). Selective pressures on the once and future contents of ethnic stereotypes: Effects of the communicability of traits. *Journal of Personality and Social Psychology*, *82*(6), 861.

Smart, S., & Waldfogel, J. (1996). *A citation-based test for discrimination at economics and finance journals*.

Sterling, J., Jost, J. T., & Bonneau, R. (2020). Political psycholinguistics: A comprehensive analysis of the language habits of liberal and conservative social media users. *Journal of Personality and Social Psychology*, *118*(4), 805.

Stremersch, S., & Verhoef, P. C. (2005). Globalization of authorship in the marketing discipline:

# Style, Content, and the Success of Ideas

## Supplemental Materials

## Writing Styles Associated with Impact

**Method**

***Segmenting Articles.*** Academic articles are usually divided into three segments. First, there is a front end, which includes an introduction, literature review, and other details that occur before authors introduce what they actually did. Second, there is a middle segment in which authors introduce their contribution, such as an experiment or analysis of field data. Third, there is a closing segment which may include a general discussion, limitations and implications of the work, and directions for future research.

To identify these three segments in each paper, we started with a manual approach. Research assistants analyzed 10 random articles from each journal and identified patterns for section names. Seven journals had consistent patterns such as starting with a roman number (e.g., I or II) or using a capital word (e.g., CONCLUSION) so sections were separated accordingly.

For the rest of the articles, supervised machine learning along with rule-based algorithms were used. Ideally one would use section titles and predict the segment label for each section, but such information was not available for many articles in our dataset. Some articles do not have clear section headers, or distinctive information (such as capital letters or bolded font) that allows section headers to be identified. Consequently, a mix of machine learning and rule-based algorithm was employed.

First, to create a training set, research assistants manually coded 700 articles based on segment breaks. Articles were divided into 5-sentence chunks and each chunk was assigned a label (i.e., 1, 2, or 3, corresponding to front end, middle, or end) based on the segment it fell into. A support vector machine (SVM) model was trained on the data using tf-idf features.

Second, the trained model was applied on the remaining set of articles, predicting a label for each 5-sentence chunk. The model achieves 78% accuracy on identifying labels for 5-sentence chunks.

To turn these labels into predictions of section breaks, and clean up any remaining issues, several rule-based steps were then applied to the output. Articles start with the front end, followed by the middle and end with the closing section. Articles cannot start with a 5-sentence chunk labeled as 3, for example, or end with a chunk labeled as 1. Further, segments should be continuous; if eight chunks out of nine consecutive chunks are labeled as 2, but the middle chunk is labeled as 1, the middle chunk should most probably be labeled as a 2 as well. Consequently, the following steps were applied on the machine learning output.

First, the first three and last three 5-sentence chunks were labeled as segment 1 and segment 3 respectively. Second, if any of the chunks start with the word "conclusion" or "discussion" (meaning that these words appeared at the beginning of a sentence), they and any chunk after were labeled as segment 3. Third, if surrounding chunks all have the same label, but that label is different than the middle chunk, the middle chunk's label was replaced with the surrounding chunks' label.

In most cases, the middle section of articles is much longer than the other two sections, so the machine learning model might be biased toward predicting label 2 for most chunks. Further, even after applying the rules discussed so far, the labels for the three segments might still be intertwined.

Consequently, two more rules were applied. First, given that the longest sequence of chunks with the same label is likely labeled correctly, the longest sequence of chunks for each label was identified. If those sequences were in the expected order (i.e., 1 before 2 before 3), chunks before the longest sequence of 1 were labeled as 1 and chunks after the longest sequence of 3 were labeled as 3.

Second, the last chunk of segment 1 and the first chunk of segment 3 were identified. To do this, the list of all sequences labeled 1 was identified. These sequences could be one or more chunks apart. Starting from the beginning of an article, if the next sequence labeled 1 is one or two chunks apart, the chunks between them (at most two) were labeled as 1. The list of all sequences labeled 3 was identified similarly. Starting from the end of an article, if the previous sequence labeled 3 is one or two chunks apart, the chunks between them (at most two) were labeled 3. What remains between the last chunk in the now contiguous sequence of 1 chunks and the first chunk in the now contiguous sequence of 3 chunks was labeled as 2.

After applying these steps, each article starts with a sequence of chunks labeled as 1 (combined as segment 1), followed by a sequence of chunks labeled as 2 (combined as segment 2) and ends with a sequence of chunks labeled as 3 (combined as segment 3).

**Ancillary Analyses**

*Simplicity.* Looking beyond function words, standard measures of readability (Flesch, 1948) find similar effects. The readability score is calculated based on the number of words per sentence and number of syllables per words with some predefined factors. Higher scores indicate texts that are easier to read. Papers with more readable front ends are cited more ($b = 0.002$, $p < .05$) but readability has a negative effect in the middle section of the paper ($b = -0.002$, $p < .01$).

Readability considers both word and sentence length, and indeed, front ends with less complex words (i.e., longer than six letters) and shorter sentences are cited more ($b = -0.007, p < .01; b = -0.003, p = .06$, respectively).

**Table A1. Descriptive Statistics**

|  | Citation Count | Word Count | Personal Pronouns | Impersonal Pronouns | Articles | Prepositions | Auxiliary Verbs | Adverbs | Conjunctions | Negations | Quantifiers |
|---|---|---|---|---|---|---|---|---|---|---|---|
| Mean | 273.20 | 11723.42 | 1.28% | 3.17% | 7.87% | 13.73% | 4.58% | 2.32% | 5.09% | 0.72% | 2.20% |
| SD | 582.62 | 5749.53 | 0.73% | 0.84% | 1.54% | 1.50% | 1.01% | 0.65% | 0.80% | 0.28% | 0.69% |

**Table A2. Predicted R2 for Style Words and Citations**

|  | Negative Binomial | | | |
|---|---|---|---|---|
|  | Baseline | +Style | Baseline + Content | +Style |
| *Predicted $R^2$* | 0.143 | 0.159*** | 0.217 | 0.224* |
| Style Features |  | yes |  | yes |
| Content Controls |  |  |  |  |
|   LDA Topics |  |  | yes | yes |
| Non-Language Controls |  |  |  |  |
|   Publication Year | yes | yes | yes | yes |
|   Journal | yes | yes | yes | yes |
|   Article Length | yes | yes | yes | yes |
|   Abstract Length | yes | yes | yes | yes |
|   Title Length | yes | yes | yes | yes |
|   Article Order | yes | yes | yes | yes |
|   Num Authors | yes | yes | yes | yes |
|   Author Gender | yes | yes | yes | yes |
|   Num References | yes | yes | yes | yes |
|   Article Type | yes | yes | yes | yes |
| Observations | 28774 | 28774 | 28774 | 28774 |

*Note.* ∗∗∗$p < .001$, ∗∗$p < .01$, ∗$p < .05$, values compared adding style features in each of the three model comparisons.

STYLE, CONTENT, AND THE SUCCESS OF IDEAS											39**Table A3. Full Details for Models with All Robustness Check Factors**

|  | Baseline | Baseline +Style |
|---|---|---|
| *Adjusted $R^2$* | **0.146** | **0.187*** |
| Style Features | yes | yes |
| Non-Language Controls | | |
|   Publication Year | yes | yes |
|   Journal | yes | yes |
|   Article Length | yes | yes |
|   Abstract Length | yes | yes |
|   Title Length | yes | yes |
|   Article Order | yes | yes |
|   Num Authors | yes | yes |
|   Author Gender | yes | yes |
|   Num References | yes | yes |
|   Article Type | yes | yes |
| Content Controls | | |
|   LDA | | |
|   LIWC Psychological Process | yes | yes |
| Other Robustness Checks | | |
|   Readability | yes | yes |
|   US Author Ratio | yes | yes |
|   Maximum Citation Count | yes | yes |
|   Highest Institution Ranking | yes | yes |
| Observations | 13249 | 13249 |

*Note.* ∗∗∗$p < .001$, ∗∗$p < .01$, ∗$p < .05$, values compared adding style features in each of the three model comparisons.

# References

Flesch, R. (1948). A new readability yardstick. *Journal of Applied Psychology*, *32*(3), 221.